\documentclass[conference]{IEEEtran}
\IEEEoverridecommandlockouts
\usepackage{cite}
\usepackage{amsmath,amssymb,amsfonts}
\usepackage{algorithmic}
\usepackage{graphicx}
\usepackage{subcaption}
\usepackage{textcomp}
\usepackage{comment}
\usepackage[none]{hyphenat}

\usepackage{xcolor}
\def\BibTeX{{\rm B\kern-.05em{\sc i\kern-.025em b}\kern-.08em
    T\kern-.1667em\lower.7ex\hbox{E}\kern-.125emX}}

\title{AI-Driven Multi-Hop Relay Selection for Smart Urban NR-V2X Networks via Learning-to-Optimize Graph Neural Networks\\
{\footnotesize %\textsuperscript{*}Note: Sub-titles are not captured for https://ieeexplore.ieee.org  and}
\thanks{}
}

\author{\IEEEauthorblockN{
Giambattista Amati, Federica Mangiatordi, Simone Angelini, Emiliano Pallotti,Pierpaolo Salvo}
\textit{Fondazione Ugo Bordoni}\\
Rome, Italy \\
Emails: \{gamati, fmangiatordi, sangelini, epallotti, psalvo\}@fub.it}

}
\begin{document}
\maketitle
\begin{abstract}
Reliable and low-latency NR-V2X communications are essential for smart mobility in dense urban environments. However, limited Road-Side Unit (RSU) density, frequent non-line-of-sight conditions, and highly dynamic vehicular topologies often prevent many Connected and Automated Vehicles (CAVs) from maintaining stable single-hop connectivity. Although multi-hop relay-assisted communication can extend infrastructure coverage, selecting relay links in real time under practical flow, capacity, and connectivity constraints remains challenging.

Mixed-Integer Linear Programming (MILP) yields optimal multi-hop relay decisions, but its computational complexity scales sharply with network density, limiting real-time applicability. To address this, we propose a Learning-to-Optimise (L2O) framework based on Graph Neural Networks (GNNs) for real-time NR-V2X relay selection. Vehicular communication states are modeled as attributed graphs, where CAVs and RSUs are nodes and candidate radio links are enriched with propagation-aware features. An offline MILP oracle provides optimal supervision, while an edge-aware Graph Isomorphism Network (GINE) approximates oracle decisions with near-constant inference latency.
Experiments on large-scale urban datasets generated by an integrated SUMO–GEMV2 simulation pipeline show that the proposed approach achieves connectivity comparable to that of the MILP oracle while reducing execution time by orders of magnitude. The framework enables cost-effective enhancement of urban V2X connectivity by leveraging existing vehicular assets and supporting scalable, real-time NR-V2X operation in smart city environments.
\end{abstract}

\begin{IEEEkeywords}
B5G technology, Mixed-Integer Linear Programming (MILP), Relay-link optimisation, Vehicular communications, Learning-to-Optimise (L2O).
\end{IEEEkeywords}
\section{Introduction}
\label{sec:introduction}

Smart cities rely on intelligent transportation systems to improve road safety, traffic efficiency, and environmental sustainability. In this context, 5G-Advanced and beyond NR-V2X technologies enable advanced services such as cooperative perception, automated driving, and real-time traffic management~\cite{Pawar10767678, Dressler8449064, su12166469}. These services impose stringent reliability, latency, and scalability requirements, particularly in dense urban environments characterized by complex road layouts, frequent non-line-of-sight (NLoS) conditions, and highly dynamic vehicular topologies~\cite{etsi_ts_122186_v18, boban2018usecases, Soto1010162021}. Consequently, conventional single-hop vehicle-to-infrastructure connectivity is often insufficient to guarantee stable coverage.

Road-Side Units (RSUs) provide key access points to the network core, yet their limited density, deployment costs, and urban obstructions often prevent a significant fraction of Connected and Automated Vehicles (CAVs) from maintaining reliable single-hop connections~\cite{6226902,8406256}. Multi-hop relay-assisted communication can therefore extend infrastructure coverage by enabling cooperative forwarding among vehicles, leveraging existing vehicular assets without requiring additional fixed roadside infrastructure.

Selecting relay links in real time under practical connectivity, flow, and capacity constraints constitutes a challenging optimisation problem. Mixed-Integer Linear Programming (MILP) formulations can compute optimal multi-hop relay configurations, but their computational complexity grows rapidly with network density, limiting real-time applicability in dense urban scenarios. Conversely, lightweight heuristics such as greedy Signal-to-Noise Ratio (SNR)-based selection are computationally efficient but myopic, as they rely on local information and fail to capture global relay interactions and multi-hop dependencies.

To bridge the gap between optimality and real-time feasibility, this paper proposes a Learning-to-Optimise (L2O) framework based on Graph Neural Networks (GNNs) for NR-V2X relay selection. Vehicular communication snapshots are modelled as attributed graphs, enabling joint reasoning over network topology and radio-link characteristics. An offline MILP oracle provides optimal supervision, while an edge-aware Graph Isomorphism Network (GINE) learns to approximate oracle decisions with near-constant inference latency suitable for real-time deployment.

The main contributions of this work are summarised as follows:
\begin{itemize}
    \item We formulate the NR-V2X multi-hop relay selection problem under practical connectivity, capacity, and forwarding constraints, and use an MILP solver as an offline oracle to generate optimal training labels.
    \item We design and evaluate a Learning-to-Optimise framework based on Graph Isomorphism Networks, comparing a topology-driven GIN model with an edge-aware GINE variant that explicitly incorporates radio link features.
    \item We conduct extensive experiments on large-scale, realistic vehicular datasets and demonstrate that the proposed L2O models achieve connectivity performance close to the MILP oracle, reduce execution time by orders of magnitude compared to optimisation-based solutions, and consistently outperform greedy heuristics.
\end{itemize}
\section{Related Work and Motivation}
\label{sec:related_work}

% ——— ML for 5G/V2X networks ———
Recent research on intelligent network optimisation for 5G and beyond has extensively explored machine learning techniques for service prioritisation, outage management, network slicing, and resource allocation~\cite{10736694,10215916,naumann2019v2x}. Although these methods demonstrate the potential of data-driven control in both cellular and vehicular networks, they typically rely on tabular feature representations, localised decision policies, or coarse network abstractions that disregard the underlying communication topology. As a result, they do not directly address the problem of multi-hop relay selection in highly dynamic vehicular environments, where network topology, traffic patterns, and link conditions evolve on sub-second time scales.

% ——— Optimisation and heuristic relay selection ———
Relay-assisted V2X communications have been investigated through both optimisation-based formulations and heuristic strategies. Exact approaches based on Mixed-Integer Linear Programming (MILP) can determine optimal relay configurations that maximise connectivity under practical constraints such as flow conservation and link capacity~\cite{onireti2016cell,zhang2019milp}. However, their computational complexity grows super-linearly with the number of nodes and candidate links, restricting their applicability to offline planning or performance benchmarking rather than real-time operation in dense urban scenarios. Conversely, heuristic methods such as SNR-driven greedy relay selection offer negligible execution time but rely on purely local information and lack global awareness, often yielding suboptimal connectivity and inefficient spectrum utilisation in multi-hop settings~\cite{boban2016geometry,cheng2018relay}.

% ——— GNN and L2O for combinatorial network problems ———
More recently, Graph Neural Networks (GNNs) and Learning-to-Optimise (L2O) paradigms have emerged as promising tools for approximating combinatorial solvers in complex networking problems. Xu~et~al.~\cite{xu2019gin} introduced the Graph Isomorphism Network (GIN), establishing a theoretical link between GNN expressiveness and the Weisfeiler--Leman graph isomorphism test. Veli\v{c}kovi\'{c}~et~al.~\cite{velivckovic2018gat} proposed Graph Attention Networks (GATs), enabling adaptive neighbour weighting through attention mechanisms. At a broader methodological level, Bengio~et~al.~\cite{bengio2021l2o} provided a meta-learning perspective on learning to optimise, while Li~et~al.~\cite{li2021combinatorial} demonstrated that GNN-based models can effectively learn to solve NP-hard combinatorial optimisation problems with competitive solution quality and significantly reduced computation time.

Within the wireless networking domain, GNN-based approaches have been applied to power control, link scheduling, and resource allocation in cellular systems ~\cite{
Shen2020GraphNN, Amati2025TopologyIG, Sun2024GraphAN,icumt_AmatiMPAS25 , Dai2025ASO
}. However, these works predominantly target single-hop resource management and do not account for the relay path selection, flow conservation, and acyclicity constraints that characterise multi-hop vehicular topologies. To the best of our knowledge, no prior work has applied a supervised L2O framework based on edge-aware GNNs to real-time multi-hop relay selection in NR-V2X networks, nor has any existing study provided a systematic comparison against both optimal MILP oracles and lightweight heuristic baselines under realistic urban propagation conditions.

% ——— Positioning and motivation ———
Motivated by these gaps, this paper proposes a graph-based Learning-to-Optimise framework explicitly designed for real-time multi-hop relay selection in NR-V2X networks. By combining offline MILP-based supervision with an edge-aware Graph Isomorphism Network (GINE) that embeds radio-link features directly into the message-passing process, the proposed approach bridges the gap between optimality and real-time feasibility, enabling scalable and deployable relay selection for smart urban mobility infrastructure
.
\begin{figure}
    \centering
    \includegraphics[width=0.99\linewidth]{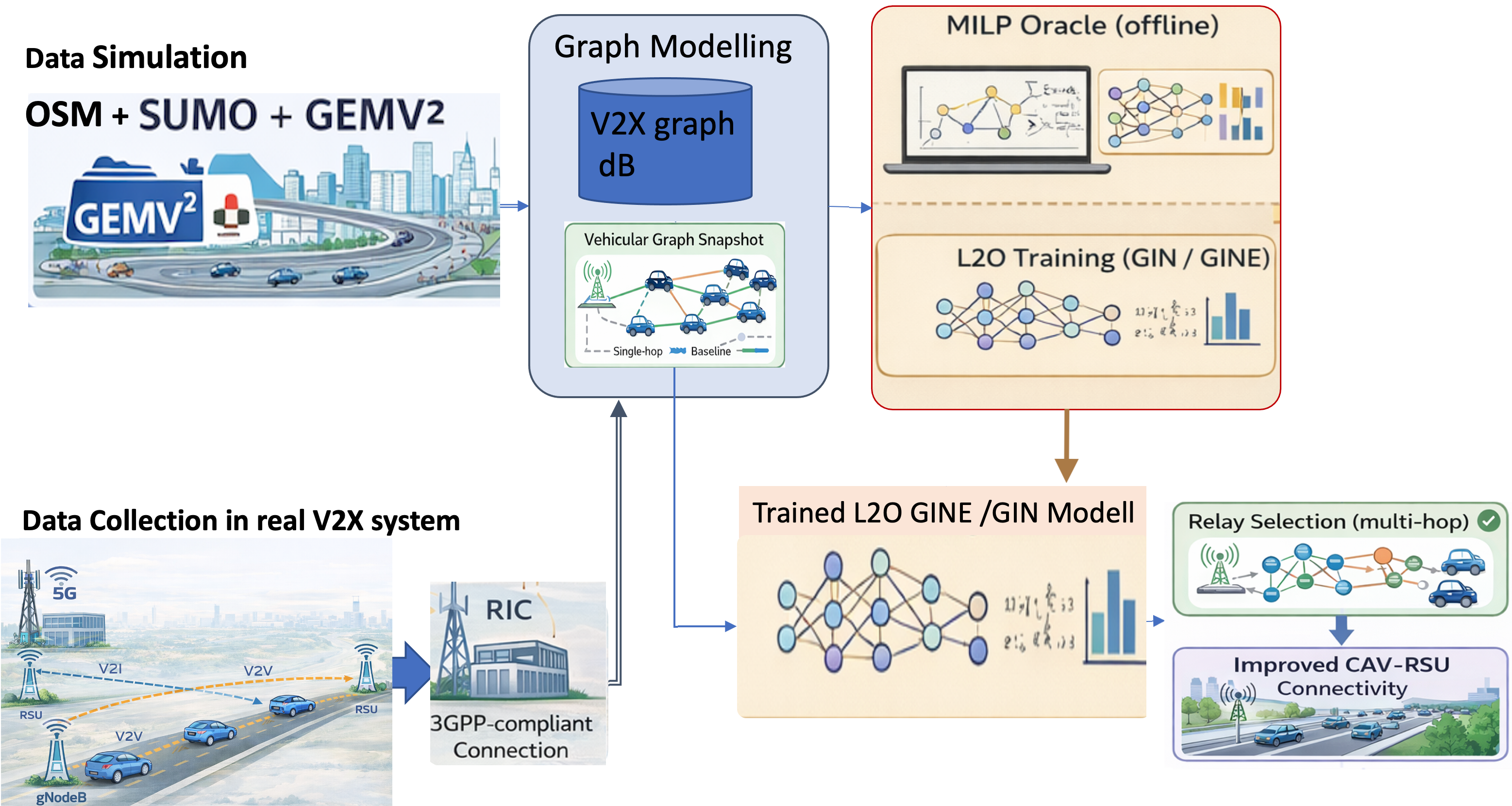}
    \caption{ GINE L2O System for NR-V2X relay selection}
    \label{fig:placeholder}
\end{figure}
\section{System Model and Problem Formulation}
\label{sec:system_model_Problem_formulation}

\subsection{System Model}

We consider a dense urban NR-V2X scenario composed of Connected and Automated Vehicles (CAVs) and
Road-Side Units (RSUs, deployed over a two-dimensional geographical area.
Vehicles are mobile and may dynamically enter or leave the considered region, whereas RSUs are
static infrastructure nodes characterised by higher transmission power and more stable connectivity.

At a given time instant, the vehicular communication network is modelled as a directed attributed
graph
\begin{equation}
\mathcal{G} = (V, E),
\end{equation}
where the node set $V$ includes both CAVs and RSUs, and the edge set $E$ represents candidate wireless
communication links.
A directed edge $(i,j) \in E$ denotes the feasibility of establishing a unicast transmission from
node $i$ to node $j$.

Each node $i \in V$ is associated with a feature vector capturing its type (CAV or RSU), spatial
position, and traffic demand when applicable.

Each edge $(i,j) \in E$ is enriched with radio-aware attributes, including the Signal-to-Noise Ratio
(SNR), achievable Shannon capacity, and link distance.
Due to their higher transmission power and elevated deployment, RSU--CAV links generally exhibit
higher SNR and capacity than CAV--CAV links.

Vehicular state information and radio link quality indicators, such as position, speed, and
SNR/SINR, are assumed to be locally measured and reported in accordance with 3GPP NR-V2X
specifications.

Time is discretized into snapshots, and the relay selection problem is solved independently at each
snapshot.
This modeling choice reflects the fast-varying nature of vehicular topologies and supports
low-latency, real-time decision making.

\subsection{Relay-Based Communication Model}
We focus on uplink or sidelink-assisted multi-hop communication, where CAVs may forward traffic either directly to an RSU or indirectly via other CAVs acting as relays. Each CAV generates a known traffic demand that must be delivered to an RSU through a valid multi-hop path.

To limit signaling overhead and hardware complexity, each CAV is constrained to activate at most one outgoing communication link. As a result, the selected relay topology forms a directed forest rooted at RSUs. CAVs that forward traffic for other vehicles act as relay nodes, while RSUs serve as traffic sinks.
\subsection{Problem Formulation}
The objective is to determine, at each snapshot, a subset of active edges
\begin{equation}
E^\star \subseteq E
\end{equation}
that maximizes network connectivity while satisfying practical communication constraints.

Specifically, the goal is to maximize the number of CAVs that are successfully connected to at least one RSU through a valid multi-hop path. This objective reflects the fundamental aim of improving coverage and service availability in dense NR-V2X deployments.

\subsubsection{Decision Variables}

Let
\begin{equation}
x_{ij} \in \{0,1\}, \quad \forall (i,j) \in E,
\end{equation}
denote a binary decision variable that equals $1$ if the directed edge $(i,j)$ is selected (active) and $0$ otherwise.

Let
\begin{equation}
y_i \in \{0,1\}, \quad \forall i \in V_{\mathrm{CAV}},
\end{equation}
represent whether CAV $i$ is successfully connected to at least one RSU.

\subsubsection{Optimization Problem}

The relay selection problem can be formulated as:
\begin{equation}
\max_{\{x_{ij}, y_i\}} \quad \sum_{i \in V_{\mathrm{CAV}}} y_i
\end{equation}

subject to the following constraints.

\paragraph{Single Outgoing Link Constraint}
Each CAV can select at most one outgoing edge:
\begin{equation}
\sum_{(i,j)\in E} x_{ij} \leq 1, 
\quad \forall i \in V_{\mathrm{CAV}}.
\end{equation}

\paragraph{Flow Conservation}
For each relay CAV, the incoming traffic flow must equal the outgoing flow:
\begin{equation}
\sum_{(k,i)\in E} f_{ki} = \sum_{(i,j)\in E} f_{ij},
\quad \forall i \in V_{\mathrm{CAV}} \setminus V_{\mathrm{RSU}},
\end{equation}
where $f_{ij}$ denotes the traffic flow on edge $(i,j)$.

\paragraph{Capacity Constraint}
The traffic forwarded over an active edge cannot exceed its achievable capacity:
\begin{equation}
f_{ij} \leq C_{ij} x_{ij},
\quad \forall (i,j)\in E,
\end{equation}
where $C_{ij}$ is the link capacity derived from the corresponding SNR.

\paragraph{Connectivity Constraint}
Each connected CAV must have a directed path to at least one RSU:
\begin{equation}
y_i \leq \sum_{j \in V_{\mathrm{RSU}}} \mathbb{I}\{ \text{path exists from } i \text{ to } j \},
\quad \forall i \in V_{\mathrm{CAV}},
\end{equation}
where $\mathbb{I}(\cdot)$ is an indicator function.

\paragraph{Loop Avoidance}
Cycles in the relay topology are not allowed, ensuring acyclic forwarding paths:
\begin{equation}
\mathcal{G}(V, E^\star) \text{ is acyclic}.
\end{equation}

\paragraph{Maximum Hop Constraint (Optional)}
To limit latency and signalling overhead, the hop count between any connected CAV and an RSU can be bounded:
\begin{equation}
\text{hop}(i, j) \leq H_{\max},
\quad \forall i \in V_{\mathrm{CAV}}, \; j \in V_{\mathrm{RSU}}.
\end{equation}

\subsection{D. Optimization and Learning-to-Optimize Perspective}
The above problem can be formulated as a Mixed-Integer Linear Program (MILP), which provides an optimal solution under the given constraints. However, the computational complexity of the MILP grows rapidly with the number of nodes and candidate edges, making it unsuitable for real-time operation in dense urban scenarios.

To overcome this limitation, we adopt a Learning-to-Optimise (L2O) approach. A Graph Neural Network is trained offline to learn the mapping between the network's graph representation and the optimal edge activation decisions produced by the MILP oracle. Once trained, the model enables fast inference, allowing near-instantaneous relay selection at runtime.
In addition to the learning-based solutions, a greedy SNR-based heuristic is considered as a lightweight baseline, enabling a comprehensive comparison between optimal, heuristic, and AI-driven approaches.

\section{Learning-to-Optimize Framework}
\label{sec:L2O_framework}

This section describes the proposed Learning-to-Optimize (L2O) framework
for real-time multi-hop relay selection in smart urban NR-V2X networks.
The approach bridges optimal but computationally expensive MILP-based
solutions and deployable, low-latency inference through Graph Neural
Networks (GNNs). A greedy SNR-based strategy is included as a lightweight
reference baseline.

\subsection{Offline MILP Oracle}

The optimisation problem defined in Section~III is solved offline via
Mixed-Integer Linear Programming (MILP) to maximise CAV--RSU connectivity
under practical constraints: (i) single outgoing link per CAV,
(ii) flow conservation, (iii) link-capacity feasibility, and
(iv) acyclicity of the relay topology.

Due to its super-linear growth with graph density, MILP is unsuitable
for real-time operation in dense smart-city scenarios. In the proposed
framework, the MILP solver is therefore used exclusively to:
\begin{itemize}
\item generate optimal edge-activation labels for supervised training,
\item provide an upper-bound reference in performance evaluation.
\end{itemize}

\subsection{GNN-Based Relay Policy (GIN and GINE)}

Each vehicular snapshot is represented as a directed attributed graph,
where nodes correspond to CAVs and RSUs and edges denote feasible
NR-V2X links.

A Graph Isomorphism Network (GIN) encoder performs message passing to
compute node embeddings capturing both local neighbourhood interactions
and global topology. For each candidate edge $(i,j)$, an
edge-classification head (MLP) combines the embeddings of nodes $i$
and $j$ and outputs a relay-activation probability.

To explicitly account for heterogeneous propagation conditions typical of dense urban deployments, we adopt an edge-aware GINE variant.
In this model, radio-level edge features (e.g., SNR, LoS/NLoS condition, distance, Shannon capacity) are directly incorporated into the message-passing process. This allows the learned policy to jointly reason over topology and link quality, improving discrimination among competing multi-hop paths.

\subsection{Runtime Inference and Feasibility Enforcement}

At runtime, relay selection is obtained through a single forward pass
of the trained GNN. Edge probabilities are thresholded to identify
candidate active links.

A lightweight feasibility refinement stage enforces structural constraints—specifically single-outgoing-link selection and directed cycle elimination—thereby producing an acyclic, RSU-rooted multi-hop forwarding graph from which valid infrastructure paths are deterministically reconstructed.

Since GNN depth is fixed and post-processing is linear in the number
of edges, the overall inference latency is near-constant and compatible
with real-time NR-V2X operation in dense smart-city environments.

\subsection{Greedy SNR Baseline}

As a reference method, a greedy SNR-based strategy is considered.
Each CAV selects the feasible outgoing link with the highest SNR,
subject to capacity and hop constraints.

While computationally efficient, this approach relies exclusively on
local decisions and ignores global relay interactions and multi-hop
dependencies. It therefore serves as a lower-bound baseline for
connectivity performance.

%%%%%%%%%%%%%%%%%%
%% Performance
%%%%%%%%%%%%%%%%%%
\section{Performance Evaluation}
\label{sec:experimental_results}

This section evaluates the proposed Learning-to-Optimize (L2O) framework
under realistic NR-V2X conditions, focusing on system-level connectivity
and real-time feasibility in heterogeneous smart-city environments.
GIN/GINE models are benchmarked against an optimal MILP oracle and a greedy
SNR-based baseline.

\subsection{Experimental Setup}
\label{subsec:exp_setup}

Vehicular mobility and radio measurements are obtained via an integrated
OpenStreetMap–SUMO–GEMV$^2$ simulation chain, which supports realistic and
reproducible system-level studies.
SUMO and GEMV$^2$ are standard tools in the V2X research community and deliver
realistic vehicle mobility traces along with geometry-aware radio
propagation models. Their joint use allows reproducible performance assessment in
diverse urban environments, covering both line-of-sight (LoS) and non-line-of-sight
(NLoS) communication conditions.

\paragraph{Urban scenarios.}
We consider three representative $500\times500\,\mathrm{m}$ areas in Rome:
\emph{Porta Pia}, \emph{EUR}, and \emph{Trastevere}. These districts exhibit
different road layouts and building densities, resulting in heterogeneous
propagation conditions and candidate-link graph structures. SUMO generates
mobility traces over the corresponding road topology, while GEMV$^2$ models
distance-dependent path loss and geometry-driven LoS/NLoS conditions.

\paragraph{Graph construction.}
At each snapshot, the network is represented as a directed attributed graph
$G=(V,E)$, where nodes correspond to CAVs and RSUs and directed edges denote
feasible NR-V2X links above a predefined SNR threshold. Node features encode
node type, 2D position, and CAV traffic demand. Edge features include SNR,
LoS/NLoS indicators, link distance, and achievable Shannon capacity.
Each snapshot is annotated offline using the MILP oracle in
Section~\ref{sec:L2O_framework}.

\paragraph{Training protocol.}
Unless otherwise stated, models are trained on the union of the three districts
(multi-area training). To assess domain shift, we additionally report a
cross-area evaluation where the model is trained on one district and tested on
the other two (Section~\ref{subsec:cross_area}).

\paragraph{Implementation details.}
We use a 3-layer GINE encoder with 256-dimensional hidden embeddings and sum
aggregation. The edge-classification head is a 2-layer MLP fed with the
concatenation of source and destination embeddings (and edge features when
applicable). Training uses weighted binary cross-entropy to mitigate class
imbalance and Adam with learning rate $10^{-3}$, up to 200 epochs with early
stopping. Data are split into 70\%/20\%/10\% train/validation/test
(seed=42). The dataset comprises \textit{[286,602]} snapshots balanced across
the three districts (\textit{Porta-Pia, Eur, Trastevere} each).

\paragraph{Execution platform.}
Training is performed on an NVIDIA RTX A5000 GPU. Runtime latency is measured on
a multicore Intel i9 CPU (128\,GB DDR5 RAM) to reflect deployment-oriented
constraints.

\subsection{Evaluation Metrics}
\label{subsec:metrics}

We report: \textit{Connectivity} (fraction of CAVs connected to at least one RSU
via single- or multi-hop paths), \textit{Connectivity Gain }(relative increase
over a single-hop V2I baseline), and \textit{Execution Latency} (time per
snapshot, including post-processing).

\subsection{Connectivity Performance}
\label{subsec:connectivity}

Multi-hop relaying significantly improves connectivity compared to single-hop
V2I. Across the evaluated dataset, the MILP oracle achieves an average
connectivity gain of $12.3\%$ (upper bound under the considered constraints).
The proposed GINE-based L2O recovers most of this gain, achieving
$11.3\%$ on average with real-time latency. The topology-only GIN variant
achieves $10.78\%$, highlighting the benefit of incorporating edge-level radio
information.

Compared to greedy SNR selection, L2O connects a larger fraction of CAVs,
especially in dense scenarios where local decisions fail to capture multi-hop
dependencies and global relay interactions. Gains are heterogeneous: more than
$9.5\%$ of snapshots exceed $25.8\%$ gain and $5\%$ exceed $40.97\%$, indicating
that relaying is particularly beneficial in challenging urban topologies.

\subsection{Runtime and Scalability Analysis}
\label{subsec:runtime}
Execution latency is critical in NR-V2X relay selection. The MILP solver shows a
strong dependency on graph density, ranging from a few milliseconds in sparse
graphs to several hundreds of milliseconds in dense topologies, which confines
MILP to offline supervision. In contrast, inference with GIN/GINE stays consistently within a few milliseconds and exhibits very low variance across snapshots, thereby confirming its suitability for real-time use. The results in Table \ref{tab:speedup_milp} indicate that L2O delivers orders-of-magnitude speedups compared to MILP, while still maintaining strong connectivity performance (Section~\ref{subsec:connectivity}).
Figure~\ref{fig:placeholder} further illustrates execution time as a
function of graph complexity.

\begin{figure}[h]
    \centering
    \includegraphics[width=0.95\linewidth]{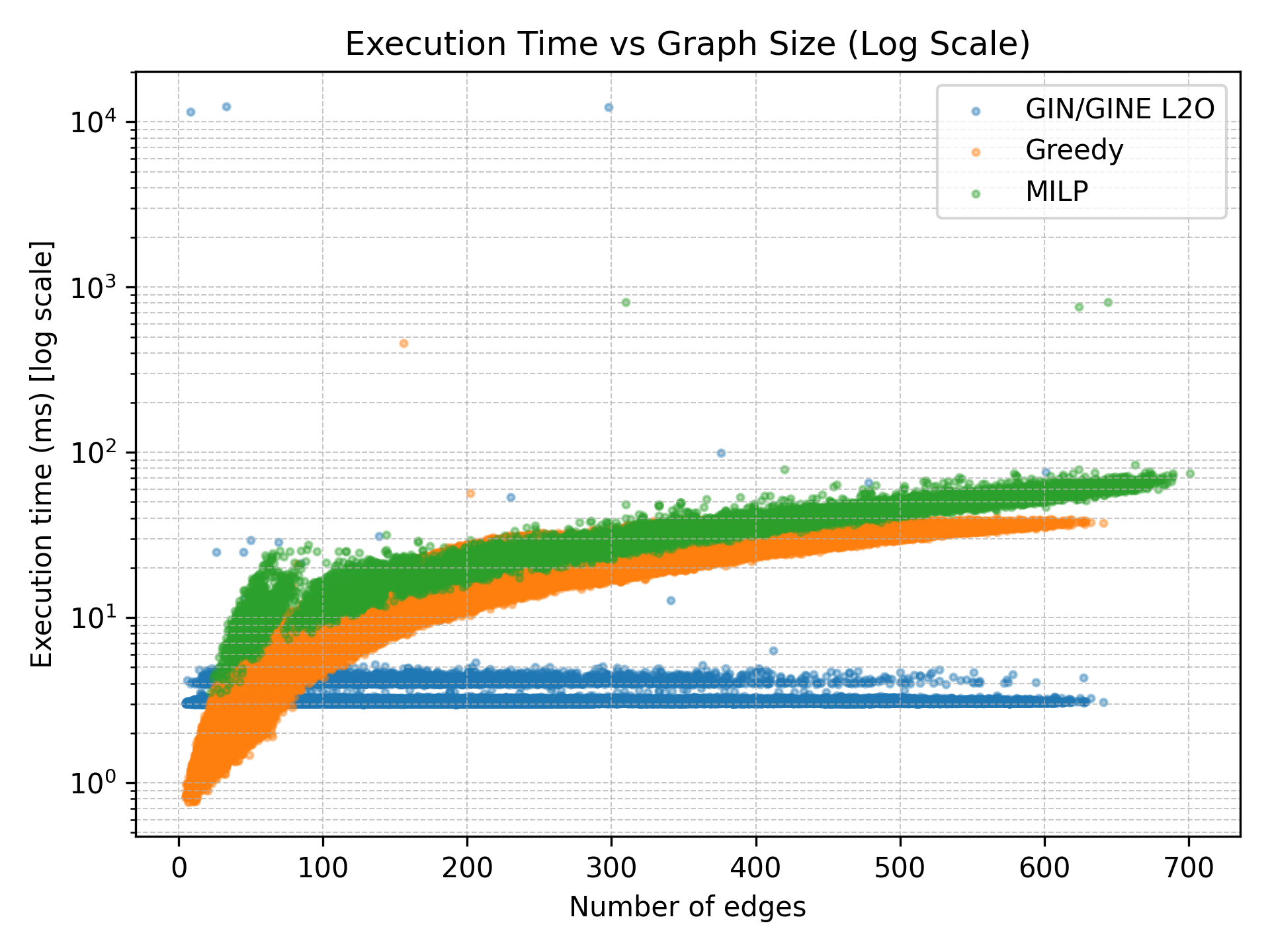}
    \caption{Execution Time vs graph complexity }
    \label{fig:placeholder}
\end{figure}

\begin{table}[th]
\centering \scriptsize
\caption{Speed-Up with Respect to MILP Across Traffic Regimes}
\begin{tabular}{lcc}
\hline
Scenario & Edge Range & Speed-Up (MILP / L2O) \\
\hline
Suburban & $<100$ edges & $\geq 10\times$ \\
Urban-Medium & $100$--$400$ edges & $15\times$--$20\times$ \\
Urban-Stress & $>400$ edges & $\geq 12\times$ \\
Worst-Case & --- & $>100\times$ \\
\hline
\end{tabular}
\label{tab:speedup_milp}
\end{table}

% (Optional) Keep ONE table only if space is tight; otherwise keep both.
% \input{...} if you prefer.

\subsection{Cross-Area Robustness Analysis}
\label{subsec:cross_area}
Smart-city deployments are inherently heterogeneous, and relay-selection policies
trained on a specific district may experience performance degradation when applied
to morphologically different areas.
To quantify urban domain shift, we evaluate cross-area generalization by training
on one district (source) and testing on the remaining two (targets), keeping the
MILP oracle and graph-construction pipeline unchanged.

\begin{table}[th]
\centering
\caption{Cross-Area Connectivity Results wi.}
\begin{tabular}{l l c c}
\hline
\textbf{Training} & \textbf{Testing} & \textbf{Gain (\%)} & \textbf{Rel. Drop (\%)} \\
\hline
Porta Pia & Porta Pia & 11.6 & -- \\
Porta Pia & EUR & 9.8 & 15.5 \\
Porta Pia & Trastevere & 8.9 & 23.3 \\
All Areas & All Areas & 11.3 & 2.6 \\
\hline
\end{tabular}
\end{table}

Results confirm measurable domain shift across districts: training on a single
area reduces connectivity gains when evaluated on unseen morphologies. Multi-area
training mitigates this degradation, yielding a more robust urban-general model
for scalable smart-city deployment. These results also provide an initial indication of the model's ability to generalise across heterogeneous urban morphologies, partially mitigating the risk of overfitting to a specific deployment environment.

\subsection{Discussion}
\label{subsec:discussion}
Results highlight that (i) multi-hop relaying is essential to improve connectivity
in dense urban NR-V2X scenarios; (ii) L2O enables near-oracle performance with
real-time latency; and (iii) edge-aware GINE consistently improves over topology-only
GIN in heterogeneous propagation environments. Cross-area evaluation further shows
that multi-area training mitigates urban domain shift and supports scalable
deployment across heterogeneous city districts.

Although the proposed framework achieves strong results on extensive urban
datasets generated through a realistic SUMO--GEMV$^2$ simulation pipeline,
its current evaluation remains simulation-based and has not yet been validated
through field-measured NR-V2X datasets or large-scale experimental testbeds.
Furthermore, the present study focuses on nominal operating conditions and does
not explicitly address adversarial communication environments, abrupt topology
changes, sensor failures, or malicious relay behaviours. Nevertheless, the
cross-area analysis provides an initial indication of the model's ability to
generalize across heterogeneous urban morphologies. Evaluating robustness under
more challenging conditions and validating the framework using real vehicular
traces and operational NR-V2X infrastructures constitute important directions
for future research.

\subsection{City-Aware Performance Analysis in Urban Areas}

To assess the practical impact of the proposed Learning-to-Optimize (L2O) framework,
we analyze 286,602 graph snapshots generated within a 500$\times$500\,m area
(0.25 km$^2$), corresponding to realistic vehicular densities observed in suburban, urban-medium, and urban-stress traffic regimes.
Vehicular density regimes are classified by using graph sparsity (measured as the number of edges) as:
\begin{itemize}
\item \textbf{Suburban regime}: sparse networks ($<100$ edges), with an average of 94 CAV.
\item \textbf{Urban-medium regime}: moderately dense networks (100--400 edges), with an average of 162 CAV.
\item \textbf{Urban-stress regime}: highly dense networks ($>400$ edges), with an average of 229 CAV.
\end{itemize}

\begin{figure}
    \centering
    \includegraphics[width=0.95\linewidth]{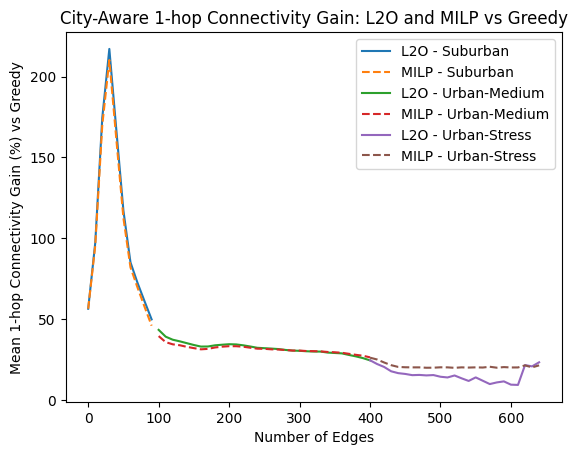}
    \caption{Connectivity Gain}
    \label{fig:placeholderfn}
\end{figure}
\paragraph{Suburban regime.}
In sparse conditions, L2O provides a median connectivity gain of 83.3\%
(mean 120.8\%), demonstrating its ability to exploit scarce relay opportunities.
In such regimes, greedy strategies suffer from myopic decisions, whereas
L2O captures global structural information.

\paragraph{Urban-medium regime.}
This represents the most realistic metropolitan operating point.
Here, L2O achieves a median gain of 33.3\%, with improvements observed in
99.4\% of graph snapshots.
This indicates a structurally robust advantage in typical urban deployments.

\paragraph{Urban-stress regime.}
Under highly dense traffic conditions, where connectivity redundancy is
naturally higher, L2O still provides a median improvement of 19.8\%.
Although the margin reduces, positive gain is maintained in 93.4\% of cases.

Overall, across all regimes, L2O improves 1-hop connectivity in more than
97\% of the evaluated snapshots, confirming that learning-based relay
selection consistently outperforms local greedy heuristics in realistic
urban vehicular environments.

%%%%%%%%%%%%%%
\section{Conclusion}
\label{sec:conclusion}
This paper investigated the problem of real-time relay selection in dense urban NR-V2X networks and proposed a Learning-to-Optimize framework based on Graph Neural Networks. By leveraging an offline MILP oracle to supervise training, the proposed approach enables fast, scalable inference while preserving most of the connectivity gains achievable by exact optimization.

Through extensive experiments on large-scale and realistic vehicular graphs, we showed that multi-hop relaying significantly improves CAV-to-RSU connectivity compared to single-hop strategies. The proposed L2O models, particularly the edge-aware GINE variant, closely approximate the MILP oracle in terms of connectivity while reducing execution time by orders of magnitude. Compared to greedy SNR-based heuristics, the learning-based approach consistently delivers higher connectivity, especially in dense and heterogeneous scenarios.

The results highlight that Graph Neural Network-based Learning-to-Optimize techniques represent a practical and effective solution for intelligent NR-V2X control, successfully balancing optimality and real-time feasibility. The proposed framework is compatible with deployment-oriented constraints and can be integrated into future 5G-Advanced and beyond vehicular networking architectures.

Future work will focus on extending the framework toward Spatio-Temporal Graph
Neural Networks (ST-GNNs) and Graph Transformer architectures capable of jointly
capturing vehicular mobility dynamics, topology evolution, and time-varying
radio conditions. Additional research will investigate validation using
real-world vehicular traces and experimental NR-V2X deployments, as well as
multi-agent and reinforcement learning approaches for decentralised relay
selection and sidelink-based communications.

\section*{Acknowledgement} %\textsuperscript{*}Note: Sub-titles are not captured for https://ieeexplore.ieee.org  and
The work has been carried out in the framework of the Spectrum Sharing project between the Ministry of Enterprises and Made in Italy (MIMIT) and Fondazione Ugo Bordoni.

%%%%%%%%%%

\bibliographystyle{IEEEtran}
\bibliography{References.bib}

@ARTICLE{Pawar10767678,
  author={Pawar, Vaishali and Zade, Nilima and Vora, Deepali and Khairnar, Vaishali and Oliveira, Aurenice and Kotecha, Ketan and Kulkarni, Ambarish},
  journal={IEEE Access},
  title={Intelligent Transportation System With 5G Vehicle-to-Everything (V2X): Architectures, Vehicular Use Cases, Emergency Vehicles, Current Challenges, and Future Directions},
  year={2024},
  volume={12},
  pages={183937-183960},
  doi={10.1109/ACCESS.2024.3506815}
}

@ARTICLE{Dressler8449064,
  author={Dressler, Falko and Klingler, Florian and Segata, Michele and Cigno, Renato Lo},
  journal={Proceedings of the IEEE},
  title={Cooperative Driving and the Tactile Internet},
  year={2019},
  volume={107},
  number={2},
  pages={436-446},
  doi={10.1109/JPROC.2018.2863026}
}

@Article{su12166469,
  AUTHOR = {Guevara, Leonardo and Auat Cheein, Fernando},
  TITLE = {The Role of 5G Technologies: Challenges in Smart Cities and Intelligent Transportation Systems},
  JOURNAL = {Sustainability},
  VOLUME = {12},
  YEAR = {2020},
  NUMBER = {16},
  ARTICLE-NUMBER = {6469},
  ISSN = {2071-1050},
  DOI = {10.3390/su12166469}
}

@techreport{etsi_ts_122186_v18,
  author       = {{ETSI}},
  title        = {{5G; Service Requirements for Enhanced V2X Scenarios}},
  institution  = {European Telecommunications Standards Institute},
  number       = {ETSI TS 122 186 V18.0.1},
  type         = {Technical Specification},
  year         = {2024},
  note         = {Release 18, May 2024}
}

@article{boban2018usecases,
  author={Boban, Mate and et al.},
  title={Use Cases, Requirements, and Design Considerations for 5G V2X},
  journal={IEEE Communications Magazine},
  year={2018}
}

@article{Soto1010162021,
  author = {Soto, Ignacio and Calderon, Maria and Amador, Oscar and Urue\~{n}a, Manuel},
  title = {A survey on road safety and traffic efficiency vehicular applications based on C-V2X technologies},
  year = {2022},
  issue_date = {Jan 2022},
  publisher = {Elsevier Science Publishers B. V.},
  address = {NLD},
  volume = {33},
  number = {C},
  issn = {2214-2096},
  doi = {10.1016/j.vehcom.2021.100428},
  journal = {Veh. Commun.},
  month = jan,
  numpages = {22}
}

@ARTICLE{6226902,
  author={Wu, Tsung-Jung and Liao, Wanjiun and Chang, Chung-Ju},
  journal={IEEE Transactions on Communications},
  title={A Cost-Effective Strategy for Road-Side Unit Placement in Vehicular Networks},
  year={2012},
  volume={60},
  number={8},
  pages={2295-2303},
  doi={10.1109/TCOMM.2012.062512.100550}
}

@INPROCEEDINGS{8406256,
  author={Premsankar, Gopika and Ghaddar, Bissan and Di Francesco, Mario and Verago, Rudi},
  booktitle={NOMS 2018 - 2018 IEEE/IFIP Network Operations and Management Symposium},
  title={Efficient placement of edge computing devices for vehicular applications in smart cities},
  year={2018},
  pages={1-9},
  doi={10.1109/NOMS.2018.8406256}
}

@INPROCEEDINGS{10736694,
  author={Amati, Giambattista and Angelini, Simone and Mangiatordi, Federica and Pallotti, Emiliano and Salvo, Pierpaolo},
  booktitle={2024 AEIT International Annual Conference (AEIT)},
  title={A Massive Clustering Algorithm for Fast Radio Resource Allocation to Distributed Units in O-RAN Networks},
  year={2024},
  pages={1-6},
  doi={10.23919/AEIT63317.2024.10736694}
}

@INPROCEEDINGS{10215916,
  author    = {Sokmaz, {\"O}mer and Yazar, Ahmet},
  booktitle = {2023 International Conference on Smart Applications, Communications and Networking (SmartNets)},
  title     = {{C-V2X} Link Decision with Machine Learning},
  year      = {2023},
  pages     = {1--5},
  doi       = {10.1109/SmartNets58706.2023.10215916}
}

@article{naumann2019v2x,
  author={Naumann, T. and et al.},
  title={V2X Communication in 5G and Beyond},
  journal={IEEE Vehicular Technology Magazine},
  year={2019}
}

@article{onireti2016cell,
  author={Onireti, O. and Zoha, A. and Imran, A.},
  title={A Cell Outage Management Framework for Dense Heterogeneous Networks},
  journal={IEEE Transactions on Vehicular Technology},
  year={2016}
}

@article{zhang2019milp,
  author={Zhang, Y. and et al.},
  title={MILP-Based Multi-Hop Routing in Wireless Networks},
  journal={IEEE Access},
  year={2019}
}

@article{boban2016geometry,
  author={Boban, M. and et al.},
  title={Geometry-Based Vehicle-to-Vehicle Channel Modeling},
  journal={IEEE Transactions on Vehicular Technology},
  year={2016}
}

@article{cheng2018relay,
  author={Cheng, L. and et al.},
  title={Relay Selection for V2X Communications},
  journal={IEEE Access},
  year={2018}
}

@inproceedings{xu2019gin,
  author={Xu, K. and et al.},
  title={How Powerful are Graph Neural Networks?},
  booktitle={ICLR},
  year={2019}
}

@article{velivckovic2018gat,
  author={{Veli\v{c}kovi\'{c}, P. and et al.}},
  title={Graph Attention Networks},
  journal={arXiv preprint arXiv:1710.10903},
  year={2018}
}

@article{bengio2021l2o,
  author={Bengio, Y. and et al.},
  title={A Meta-Learning Perspective on Learning to Optimize},
  journal={Proceedings of the IEEE},
  year={2021}
}

@article{li2021combinatorial,
  author={Li, Y. and et al.},
  title={Learning to Solve Combinatorial Optimization Problems},
  journal={IEEE Transactions on Neural Networks and Learning Systems},
  year={2021}
}

@article{Shen2020GraphNN,
  title={Graph Neural Networks for Scalable Radio Resource Management: Architecture Design and Theoretical Analysis},
  author={Yifei Shen and Yuanming Shi and Jun Zhang and Khaled Ben Letaief},
  journal={IEEE Journal on Selected Areas in Communications},
  year={2020},
  volume={39},
  pages={101-115},
  url={https://api.semanticscholar.org/CorpusID:220525894}
}

@article{Amati2025TopologyIG,
  title={Topology Intelligence: GNN-Powered 5G V2X},
  author={Giambattista Amati and Federica Mangiatordi and Emiliano Pallotti and Simone Angelini and Pierpaolo Salvo},
  journal={2025 AEIT International Annual Conference (AEIT)},
  year={2025},
  pages={1-6},
  url={https://api.semanticscholar.org/CorpusID:282941821}
}

@article{Sun2024GraphAN,
  title={Graph Attention Network Enhanced Power Allocation for Wireless Cellular System},
  author={Qiushi Sun and He Yang and Ovanes L. Petrosian},
  journal={Informatics and Automation},
  year={2024},
  url={https://api.semanticscholar.org/CorpusID:267121734}
}

@inproceedings{icumt_AmatiMPAS25,
  author       = {Giambattista Amati and
                  Federica Mangiatordi and
                  Emiliano Pallotti and
                  Simone Angelini and
                  Pierpaolo Salvo},
  title        = {5G-V2X Topology Intelligence: {A} {GNN} Approach},
  booktitle    = {17th International Congress on Ultra Modern Telecommunications and
                  Control Systems and Workshops, {ICUMT} 2025, Florence, Italy, November
                  3-5, 2025},
  pages        = {7--12},
  publisher    = {{IEEE}},
  year         = {2025},
  url          = {https://doi.org/10.1109/ICUMT67815.2025.11268640},
  doi          = {10.1109/ICUMT67815.2025.11268640},
  bibsource    = {dblp computer science bibliography, https://dblp.org}
}

@article{Dai2025ASO,
  title={A Survey of Graph-Based Resource Management in Wireless Networks—Part I: Optimization Approaches},
  author={Yanpeng Dai and Ling Lyu and Nan Cheng and Min Sheng and Junyu Liu and Xiucheng Wang and Shuguang Cui and Lin Cai and Xuemin Sherman Shen},
  journal={IEEE Transactions on Cognitive Communications and Networking},
  year={2025},
  volume={11},
  pages={2078-2100},
  url={https://api.semanticscholar.org/CorpusID:274505837}
}

\end{document}